%% file: main.tex
\documentclass{article}
\usepackage{iclr2027_conference,times}
\usepackage{hyperref}
\usepackage{url}
\usepackage{graphicx}
\usepackage{wrapfig}
\usepackage{needspace}
\usepackage{amsmath,amssymb}
\usepackage{booktabs}
\usepackage{array}
\usepackage{multirow}
\usepackage{xcolor}
\usepackage{colortbl}
\definecolor{referenceblue}{RGB}{0,70,170}
\hypersetup{colorlinks=true,linkcolor=referenceblue,citecolor=referenceblue,urlcolor=referenceblue,pdfborder={0 0 0}}
\newcommand{\resulttablestyle}{\small\renewcommand{\arraystretch}{1.16}\setlength{\tabcolsep}{7pt}}
\usepackage{kotex}

\iclrpreprintcopy
\title{RobotUse: Allocating Computation, Context, and Decisions}
\author{Junhoo Lee$^{1}$\thanks{Equal contribution.}\quad
Injun Baek$^{2}$\footnotemark[1]\quad Seungyeon Kim$^{2}$\quad Suhyun Jeon$^{2}$\\
\bfseries Minkyu Kim$^{2}$\quad Baekseung Kim$^{2}$\quad Nojun Kwak$^{2}$\\
\normalfont $^{1}$KAIST\quad $^{2}$Seoul National University\\
\normalfont\fontsize{8.5}{10}\selectfont \texttt{junhoo.lee@kaist.ac.kr}\\
\normalfont\fontsize{8.5}{10}\selectfont \texttt{\{jjune1416,syeonkim,suhyun0606,minkyu.kim,baekseung.kim,nojunk\}@snu.ac.kr}}
\hypersetup{
  pdftitle={RobotUse: Allocating Computation, Context, and Decisions},
  pdfauthor={Junhoo Lee, Injun Baek, Seungyeon Kim, Suhyun Jeon, Minkyu Kim, Baekseung Kim, Nojun Kwak},
  pdfsubject={Preprint}
}
\begin{document}
\maketitle
\input{sec/00_abstract}
\suppressfloats[t]
\input{sec/01_intro_new}
\input{sec/00_overview}
\input{sec/02_related_work}
\input{sec/03_Problem}
\input{sec/04_Method}
\input{sec/04_results}
\input{sec/05_ablation}
\input{sec/05_context}
\input{sec/05_continual}

\input{sec/06_conclusion}
\input{sec/08_llm_statement}
\bibliography{references}
\bibliographystyle{iclr2027_conference}
\clearpage
\appendix
\setcounter{topnumber}{3}
\input{sec/06_evaluation_details}

\input{sec/07_implementation}
\end{document}

%% file: sec/00_abstract.tex
\begin{abstract}
Robot agents must connect their intended actions to observed outcomes while retaining the context needed to revise their choices over repeated attempts. Existing interfaces often leave these choices inside predefined tools or require agents to manage detailed execution code and its growing history. We introduce RobotUse, a robot agent harness that organizes computation, context, and decisions around specifying and revising physical actions. Agents visually select targets and poses, while the backend handles geometry, motion planning, and control. Subagents retain detailed interactions within each subgoal and return the information needed for subsequent decisions. Continual harnessing lets agents learn from execution by updating a persistent playbook. On RoboLab, RobotUse achieves 45\% task success, outperforming CaP-X by 6.7 percentage points while maintaining compact decision contexts and reducing reliance on predefined action abstractions. Furthermore, we show that RobotUse learns from real-world execution despite imperfect feedback and transfers what it learns to subsequent tasks. Project page is available at \url{https://robotuse-team.github.io/}.
\end{abstract}

%% file: sec/01_intro_new.tex
\section{Introduction}
\label{sec:introduction}

Everyday actions give concrete form to simple intentions: a door opened just enough to pass through, a dial turned until the sound is right. These intentions guide what to change, by how much, and when to adjust, without requiring every movement to be consciously specified.

General-purpose agent design requires deciding which responsibilities belong to model reasoning and which should be delegated. A model's ability to perform a computation does not mean that the computation should be performed through model inference. Programs can process information outside the active context, while subagents can conduct detailed interactions and return the results needed for subsequent decisions~\citep{zhang2026rlm,bengre2026context}.  Recent agent harnesses organize computation and context to support long-horizon, multi-step work and retain experience across interactions~\citep{karten2026prime,karten2026continual}. The design question is how to allocate computation and context so that the model retains the decisions that matter without repeatedly re-deriving the backend computations needed to realize them.

However, physical action requires coordinating high-dimensional, continuous motion under geometric and contact constraints. Even a simple goal entails choices about position, orientation, and approach that must change with the scene and observed outcomes. Code-based policies express these choices through programs over perception and control tools~\citep{liang2022code}, leaving the agent to connect physical intentions and failures to execution logic. On the other hand, high-level skills delegate execution but can conceal choices needed for adjustment~\citep{ahn2022saycan}. Experience can improve programs and skills~\citep{lu2026aspire,chen2026gap}, yet accumulating abstractions can make the connection between internal decisions and physical outcomes harder to inspect and repair. A robot harness must therefore delegate execution complexity while retaining the context needed to interpret outcomes and guide subsequent actions.

We propose RobotUse, an intent-based agent harness that organizes computation, context, and decisions around specifying and revising physical intent (Fig. \ref{fig:overview}). The main agent maintains task intent in language and delegates subgoals and constraints to a general-purpose subagent. The subagent grounds this intent in visual observations by selecting and refining task-relevant targets, positions, and orientations. The backend translates these visually specified choices into executable motion through perception, geometric computation, motion planning, and control. Detailed observation, selection, and recovery histories remain in the subagent's local context, while outcomes, current state, and unresolved constraints return to the main agent to guide subsequent intent and delegation. A playbook guides decomposition and delegation, and accumulates experiential guidance about what worked, under which conditions, and what to try differently. Continual harnessing updates this guidance while keeping the execution backend fixed, allowing experience to change subsequent decisions without rewriting robot control code.

Our evaluation compares task completion, inference cost, token usage, and execution time on RoboLab. We examine dependence on the provided grasp tool and track task completion across successive harness revisions. Across successive playbook refinements, accumulated experience improves task performance. Together, these studies examine how the reasoning demands of robot manipulation depend on the responsibilities assigned to the model and how execution experience can improve the guidance governing those responsibilities.

%% file: sec/00_overview.tex
\begin{figure}[t]
    \centering
    \includegraphics[width=\linewidth]{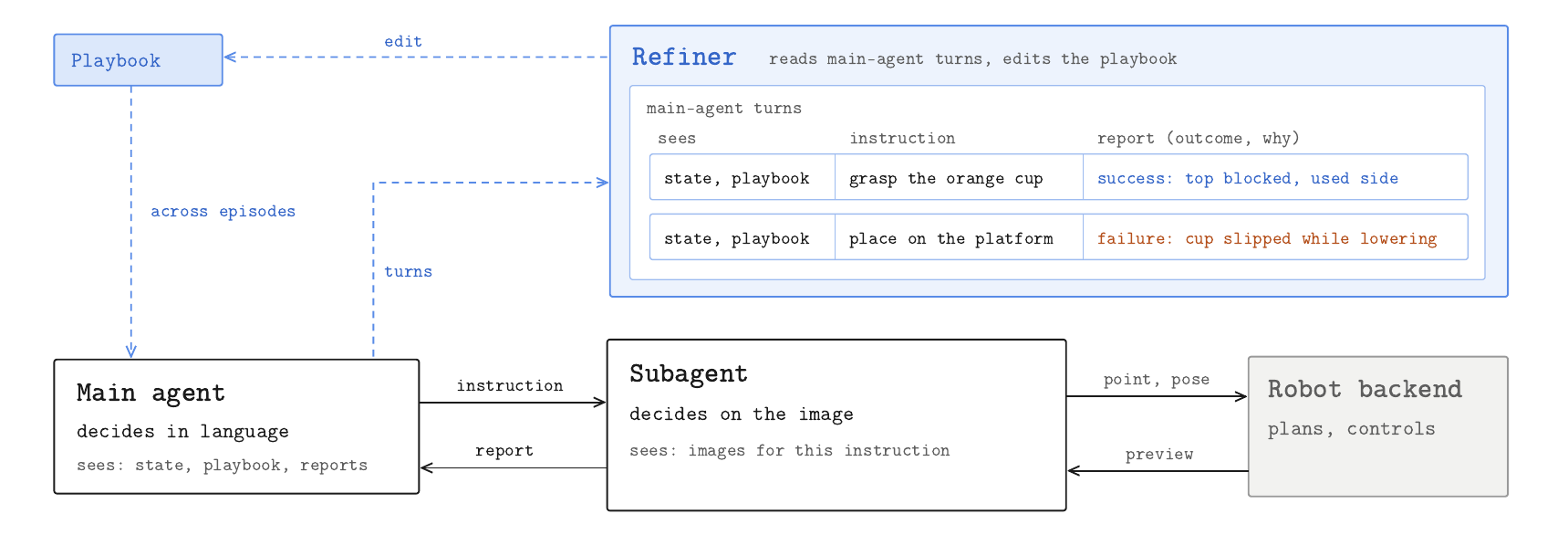}
    \caption{\textbf{RobotUse.}
    The main agent decides in language: from the current state, the playbook, and returned reports, it issues an instruction.
    The subagent decides on the image, selecting points and poses that the robot backend plans and executes, and returns a report of the outcome and its cause.
    Between episodes (dashed), a refiner reads only the main-agent turns and edits the playbook.}
    \label{fig:overview}
\end{figure}

%% file: sec/02_related_work.tex
\section{Related Work}
\label{sec:related-work}

\paragraph{Agent harnesses.}
Agent harnesses organize the computation and information through which a language model acts.
ReAct interleaves reasoning with actions and observations to revise plans during interaction~\citep{yao2023react}.
SWE-agent studies how the design of an agent's tools and environmental feedback shapes its behavior and performance~\citep{yang2024sweagent}.
Recursive Language Models treat long inputs as an external environment that the model can inspect programmatically and decompose through recursive calls~\citep{zhang2026rlm}.
Prime Agent extends this approach with persistent execution, recursive subagent sessions, and retained histories, allowing the model to construct strategies while the harness manages execution and state~\citep{karten2026prime}.
Deep Agents similarly uses delegation to isolate detailed work from the supervisor's context, with context inheritance available when a subagent needs to continue prior work~\citep{bengre2026context}.
Reflexion retains verbal feedback in episodic memory to improve decisions in subsequent trials~\citep{shinn2023reflexion}.
Continual Harness updates prompts, subagents, skills, and memory from execution trajectories~\citep{karten2026continual}.
Our work brings this allocation of computation and context to physical action.
We study the boundary at which geometric execution can be delegated while task-relevant choices remain accessible to the agent, and local interaction histories can be separated while preserving the information needed for subsequent decisions.
Experience is retained as playbook guidance over this interface with the execution backend held fixed.

\paragraph{LLMs for robotic manipulation.}
Vision-language-action models map images and instructions to robot actions, as in RT-2~\citep{brohan2023rt2}, OpenVLA~\citep{kim2025openvla}, and $\pi_0$~\citep{black2025pi0}; we include $\pi_{0.5}$~\citep{physicalintelligence2025pi05} as a representative baseline.
Early language-model robot agents separate decisions about what to do from the motor skills that carry them out.
SayCan grounds skill selection in language relevance and affordance estimates~\citep{ahn2022saycan}, while Inner Monologue uses environmental feedback to revise task plans~\citep{huang2022inner}.
PIVOT exposes continuous action choices to vision-language models by rendering candidate actions on images and refining them through iterative visual selection~\citep{nasiriany2024pivot}.
Research has also expanded the agent's responsibility to constructing and improving executable behavior.
Code as Policies generates programs that compose perception and control APIs~\citep{liang2022code}.
ASPIRE diagnoses failures from multimodal execution traces, repairs control programs, and accumulates validated improvements in a reusable skill library~\citep{lu2026aspire}.
GaP constructs graphs of perception, planning, and control modules and refines their structure and parameters through simulation rehearsal~\citep{chen2026gap}.
These approaches support flexible behavior by making executable programs and skill compositions objects of generation and revision.
Inspired by the earlier separation of decision-making and execution, we develop a harness that retains concrete physical choices at the agent interface while delegating their geometric realization to the backend.
Subgoal interactions remain in local contexts, and experience accumulates as guidance about which choices worked and how to adjust them, without modifying the underlying execution code.

%% file: sec/03_Problem.tex
\section{Problem Formulation}
\label{sec:problem}
\label{sec:state_transition}

\paragraph{Robot harness.}
A language-model agent selects commands from a robot interface using a task goal $g$ and the observations and interaction history $h_t$ provided to it. For an interface $I$ with command space $\mathcal U_I$, this decision takes the form
\begin{equation}
    u_t \sim \pi_\theta(\cdot\mid g,h_t),
    \qquad u_t\in\mathcal U_I.
    \label{eq:exposed_action_space}
\end{equation}
The execution backend interprets the command and performs the computation and control needed to realize it. We represent backend execution together with the physical environment's response as
\begin{equation}
    s_{t+1}\sim P_I(\cdot\mid s_t,u_t),
    \label{eq:execution}
\end{equation}
where $s_t$ is the physical state. The index $t$ denotes an agent interaction step, so a command can span multiple robot control steps. Observations and execution feedback contribute to the history provided for subsequent decisions.

A robot harness organizes this interaction: it determines the available commands, their realization through the backend, and the observations and history supplied to the agent. Even with the same robot capabilities, commands may expose skill invocations, executable code, individual movements, or target poses. The harness therefore allocates decisions between the agent and the execution backend.

\paragraph{Challenges.}
\emph{Delegating execution while retaining decisions.}
Robot actions require perception, geometric computation, motion planning, and control. Code-, skill-, and tool-based interfaces make these capabilities callable, reducing the agent's execution burden~\citep{liang2022code,capx,ahn2022saycan}. A high-level execution unit can also determine a task-relevant choice, such as the grasp location or approach direction (Fig.~\ref{fig:action-interface}a). Expanding that unit into perception and geometric operations can expose its implementation while still selecting a grasp through an internal scoring rule (Fig.~\ref{fig:action-interface}b). The harness must therefore make the intended choice available for selection and revision as it determines which computations to delegate.

\emph{Connecting physical outcomes to revisions.}
Actions unfold over time, and their outcomes depend on geometric and contact conditions. A failed grasp may originate in target selection, approach direction, or contact location. The unit invoked by a command can therefore differ from the choice that needs revision. The agent needs feedback that connects outcomes to these choices and an interface that lets it revise them.

\emph{Maintaining context for revision.}
Revision requires current observations, previous choices, execution results, and attempted adjustments. Repeated interaction accumulates this history within a subgoal. The next task-level decision may require the outcome, changed scene, and unresolved constraints, while further spatial correction needs the detailed sequence of local attempts. The harness must determine where these histories are retained and what information passes between decisions.

We study how to organize commands and context so that the agent can specify and revise task-relevant choices while delegating the computation required to realize them.

%% file: sec/04_Method.tex
\Needspace{4\baselineskip}
\section{RobotUse}
\label{sec:method}

\begin{figure}[t]
        \centering
        \begin{minipage}{0.94\linewidth}
        \begin{minipage}[t]{0.308\linewidth}
            \centering
            \includegraphics[width=\linewidth]{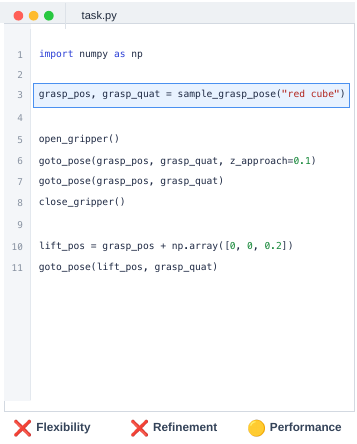}
            \par\smallskip
            {\small (a) Code-as-policy}
        \end{minipage}\hfill
        \begin{minipage}[t]{0.317\linewidth}
            \centering
            \includegraphics[width=\linewidth]{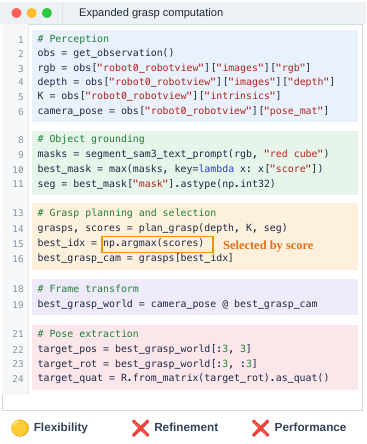}
            \par\smallskip
            {\small (b) Naive expansion}
        \end{minipage}\hfill
        \begin{minipage}[t]{0.335\linewidth}
            \centering
            \includegraphics[width=\linewidth]{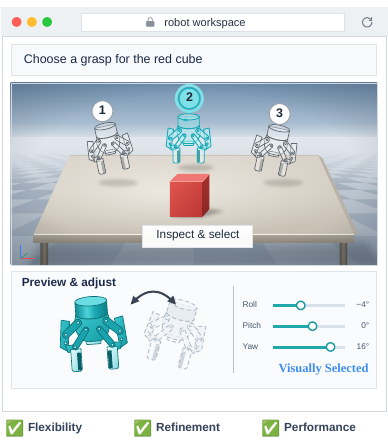}
            \par\smallskip
            {\small (c) RobotUse}
        \end{minipage}
        \end{minipage}
        \vspace{-3pt}
        \caption{\small\textbf{From code-based execution to agent-specified actions.}
        (a) Conventional code-as-policy composes high-level calls that determine grasp poses internally.
        (b) A naive expansion exposes perception, grasp generation, and coordinate transformations, but the illustrated routine still selects the highest-scoring grasp; the agent's intended grasp has no explicit selection interface.
        (c) RobotUse exposes grasp candidates for visual selection and pose previews for adjustment. The agent specifies the intended action, and the backend computes and executes the corresponding motion.}
        \label{fig:action-interface}
        \vspace{-5pt}
\end{figure}

\subsection{Design Principle}
\label{sec:design-principle}
Figure~\ref{fig:action-interface} illustrates the interface problem in the code-based manipulation~\citep{capx}. Expanding a high-level grasp call can expose its implementation while leaving the grasp choice to a score-based rule. RobotUse makes the candidate and its pose explicit objects of agent selection and revision.

RobotUse retains task-relevant physical choices at the agent interface and delegates their geometric realization to the execution backend. The agent chooses targets, locations, orientations, and adjustments; the backend performs geometric conversion, motion planning, and control. For example, grasping a cup establishes an intended outcome, while the selected grasp location and orientation specify how the gripper should engage it. These choices remain available for inspection and revision throughout execution (Fig.~\ref{fig:action-interface}c).

For spatial execution commands, we instantiate $u_t$ in Eq.~\ref{eq:exposed_action_space} as a specification $z_t$. The division between selecting a specification and realizing it is
\begingroup
\setlength{\abovedisplayskip}{4pt}
\setlength{\belowdisplayskip}{4pt}
\begin{equation}
z_t\sim\pi_\theta(\cdot\mid g_i,h_t^{(i)},P),
\qquad
\xi_t\sim B(\cdot\mid s_t,z_t).
\label{eq:spatial-specification}
\end{equation}
\endgroup
Here $g_i$ is the current subgoal, $h_t^{(i)}$ its local interaction history, and $P$ the persistent playbook. The agent selects a specification $z_t$, and the backend $B$ generates the motion $\xi_t$. Unlike $B$, $P_I$ in Eq.~\ref{eq:execution} also includes the environment's response. Pointing, candidate selection, and pose editing let the agent revise the specification using visual and diagnostic feedback.

Accurate specification can require repeated observation and adjustment. RobotUse therefore couples this decision boundary with a context boundary: detailed interactions remain within the subgoal they serve, while returned reports carry the outcomes and constraints needed for subsequent planning. The playbook guides both delegation and local execution, retaining conditional guidance across episodes.

\input{figures/qualitative/qual_combined}
\input{figures/evaluation/environments}

\subsection{Handoff Unit}
\label{sec:handoff-unit}
A handoff consists of a language-defined subgoal $g_i$, its constraints, and the relevant task context. The main agent selects the subgoal using the overall goal, the playbook, and its history $H_i$. A general-purpose subagent maintains a separate context $h_t^{(i)}$ for execution. Its returned report updates the main context:
\begin{equation}
H_{i+1}=H_i\oplus\bigl(g_i,\rho(h^{(i)})\bigr).
\label{eq:context-handoff}
\end{equation}
Here $\rho(h^{(i)})$ is the report from the local interaction history, and $\oplus$ appends the handoff and report. Both agents use the same model $\pi_\theta$ with separate contexts.

The subagent returns when the subgoal is completed or progress requires reconsidering the subgoal or its constraints. Its report describes the outcome and reason, provides the current image, and records relevant selections, revisions, and unresolved issues. These fields let the main agent assess what changed during the delegation and which constraints should guide its next action.

The local history includes the initial handoff and the observations, selections, and corrections made while pursuing it. Its report carries the findings needed to continue the task, allowing the main agent to choose another subgoal or reconsider the current one from the resulting scene. Detailed interaction histories remain in the subagent's context; the main agent accumulates the handoffs and returned reports needed to organize progress across subgoals.

\Needspace{12\baselineskip}
\subsection{Closed-Loop Interaction}
\label{sec:closed-loop-interaction}
Within a handoff, the subagent repeatedly observes the scene, specifies a spatial choice, inspects feedback, and revises or executes the choice. Pointing identifies targets or locations, candidate selection chooses among backend-generated alternatives, and pose editing adjusts position or orientation. The backend handles geometric conversion and motion generation for the selected specification. Language supplies the subgoal and its constraints, while visual representations expose the spatial choices used to realize it.

Inspection and execution provide different feedback. Previews and feasibility diagnostics help the agent examine a requested specification before motion; an infeasible request can return diagnostics without executing it. After execution, the agent uses the resulting observation and diagnostic feedback to assess progress and decide whether to revise the specification. For example, an obstructed approach can motivate a change in direction, while an incorrectly selected destination can motivate a position adjustment. The specification provides a common reference for these corrections; feedback must still be interpreted in the current scene.

The local context retains observations, selected candidates, adjustments, and execution feedback across this loop. Subsequent decisions can therefore refer to what was attempted and why it was changed. If a correction requires changing the delegated goal or constraints, the subagent returns the relevant evidence for a new task-level decision.

\paragraph{Implementation.}
We implement spatial specification as structured tool calls over calibrated RGB-D observations. Target selections are registered as references that subsequent candidate-generation calls consume. Each candidate has a reference used for inspection, adjustment, and execution. Inspecting a candidate produces a pose card with side, top, and closing-plane views of measured points and the robot gripper; an adjustment produces an updated card. The agent can translate the pose in robot-base axes and rotate it about the gripper's contact center before requesting motion. The backend resolves these requests into robot poses and returns planning or execution feedback. This makes the selected candidate the link between visual inspection and the motion request. Appendix~\ref{app:implementation} describes tool contracts, context allocation, and playbook updates.

\subsection{Continual Harness}
\label{sec:continual-harness}
RobotUse retains experience across episodes through a persistent playbook~\citep{karten2026continual}. Each round, an outer-loop refiner updates the playbook from main-agent trajectories containing delegations, returned reports, and subsequent task-level decisions. Local execution details are available to the refiner through the returned reports.

Refinement records the conditions under which particular spatial choices and recovery strategies were effective, the context required for a delegation, and the findings needed for subsequent planning. New evidence can qualify or revise earlier guidance. The resulting playbook guides both agents in delegating, selecting, and revising actions. The refiner edits the playbook while leaving model parameters and backend code unchanged; the development protocol is detailed in Appendix~\ref{app:evaluation-details}.

%% file: figures/qualitative/qual_combined.tex
\begin{figure*}[t]
\centering
\includegraphics[width=\linewidth]{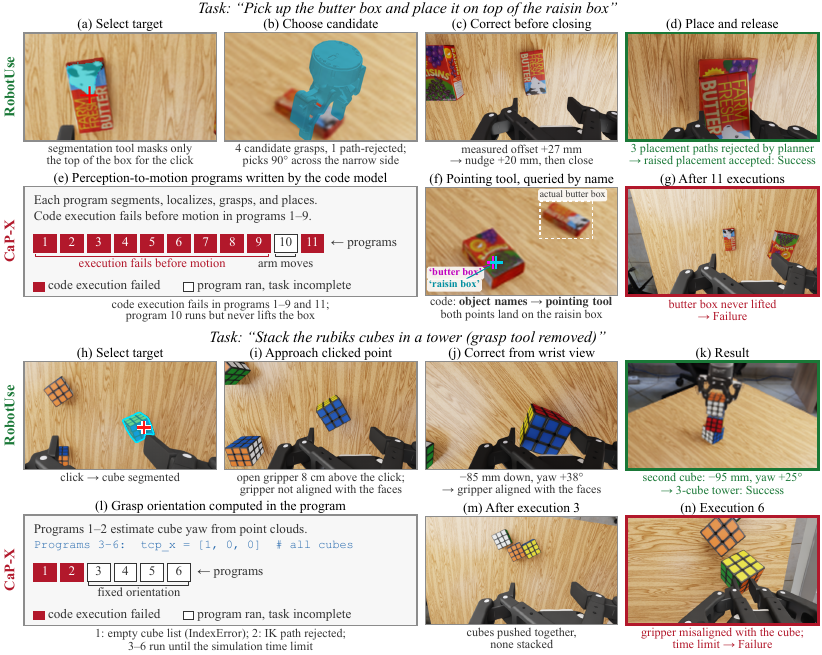}
\caption{\textbf{Qualitative comparison with CaP-X.} \mbox{(a--g)}~RobotUse checks imperfect tool results in the images and succeeds (3/3); CaP-X's code execution fails in 10 of 11 attempts, and its name-queried pointing tool marks the wrong box (0/3). \mbox{(h--n)}~Without the grasp tool, RobotUse corrects its grasp visually; CaP-X uses one orientation for every cube in programs 3--6.}
\label{fig:qual}
\end{figure*}

%% file: figures/evaluation/environments.tex
\begin{figure}[t]
\centering
\begin{minipage}[t]{0.485\linewidth}
\centering
\parbox[c][1.65cm][c]{\linewidth}{%
\includegraphics[width=0.324\linewidth,height=1.65cm,keepaspectratio]{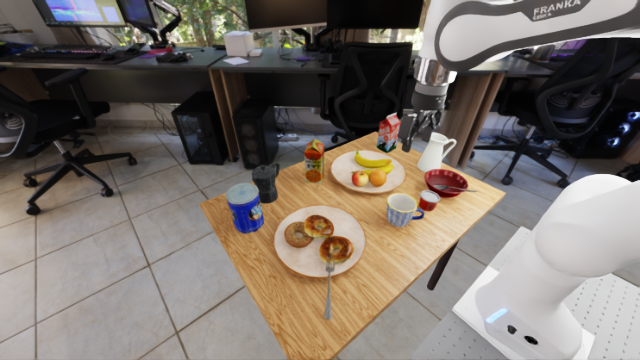}\hfill
\includegraphics[width=0.324\linewidth,height=1.65cm,keepaspectratio]{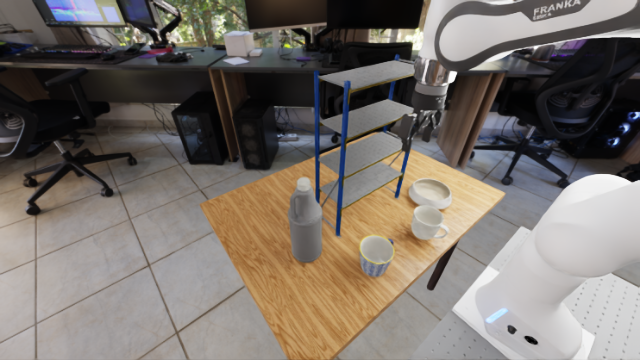}\hfill
\includegraphics[width=0.324\linewidth,height=1.65cm,keepaspectratio]{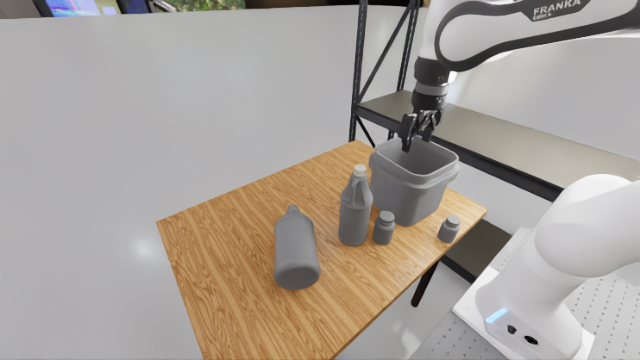}}
\par\smallskip
{\small (a) RoboLab simulation}
\end{minipage}\hfill
\begin{minipage}[t]{0.485\linewidth}
\centering
\parbox[c][1.65cm][c]{\linewidth}{%
\includegraphics[width=0.324\linewidth,height=1.65cm,keepaspectratio]{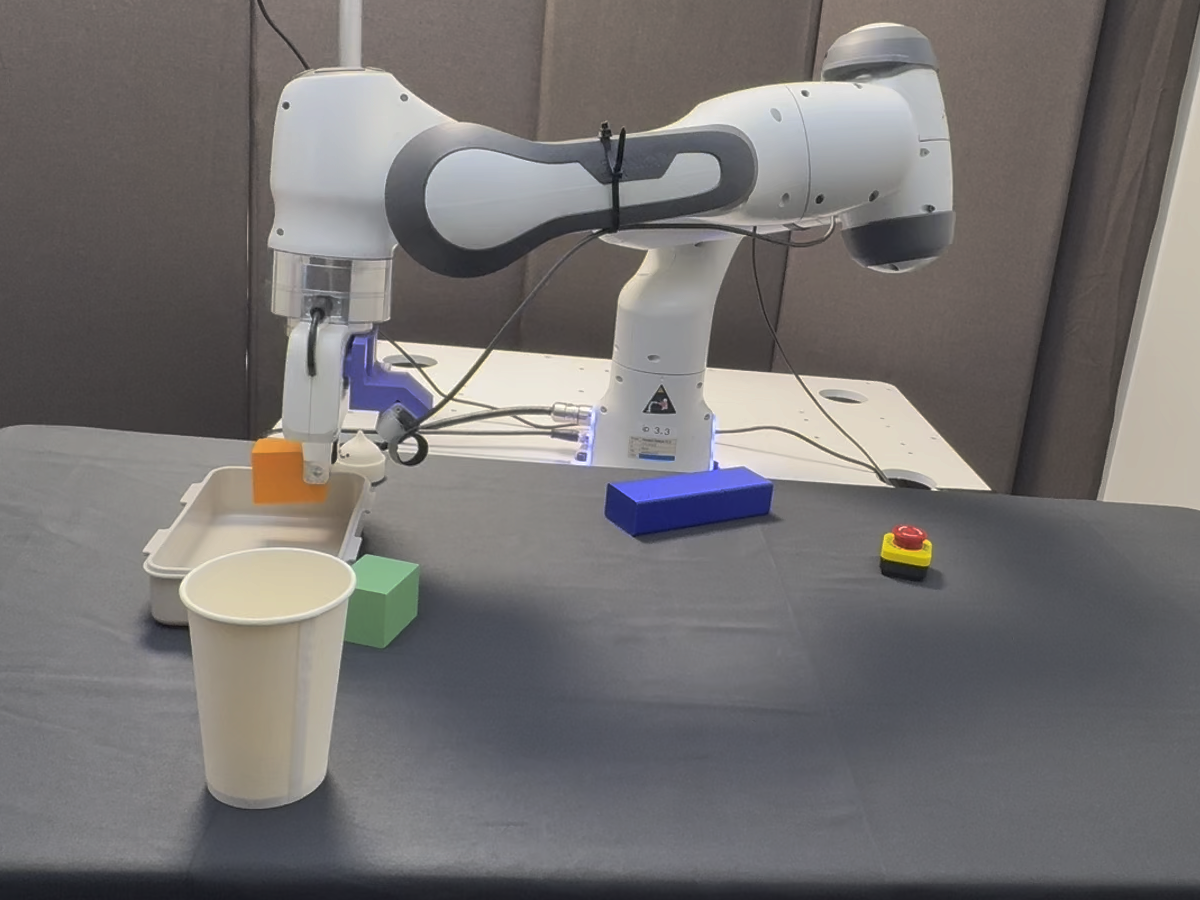}\hfill
\includegraphics[width=0.324\linewidth,height=1.65cm,keepaspectratio]{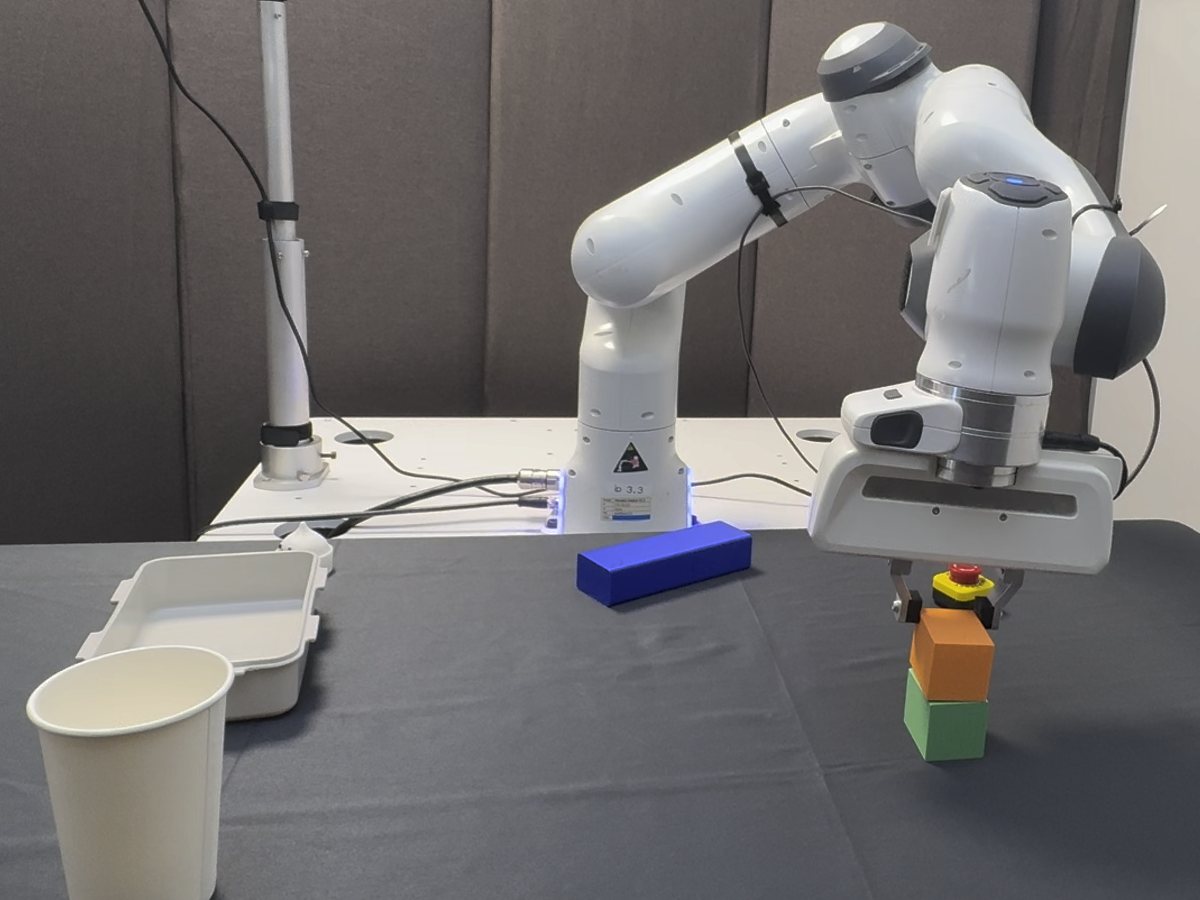}\hfill
\includegraphics[width=0.324\linewidth,height=1.65cm,keepaspectratio]{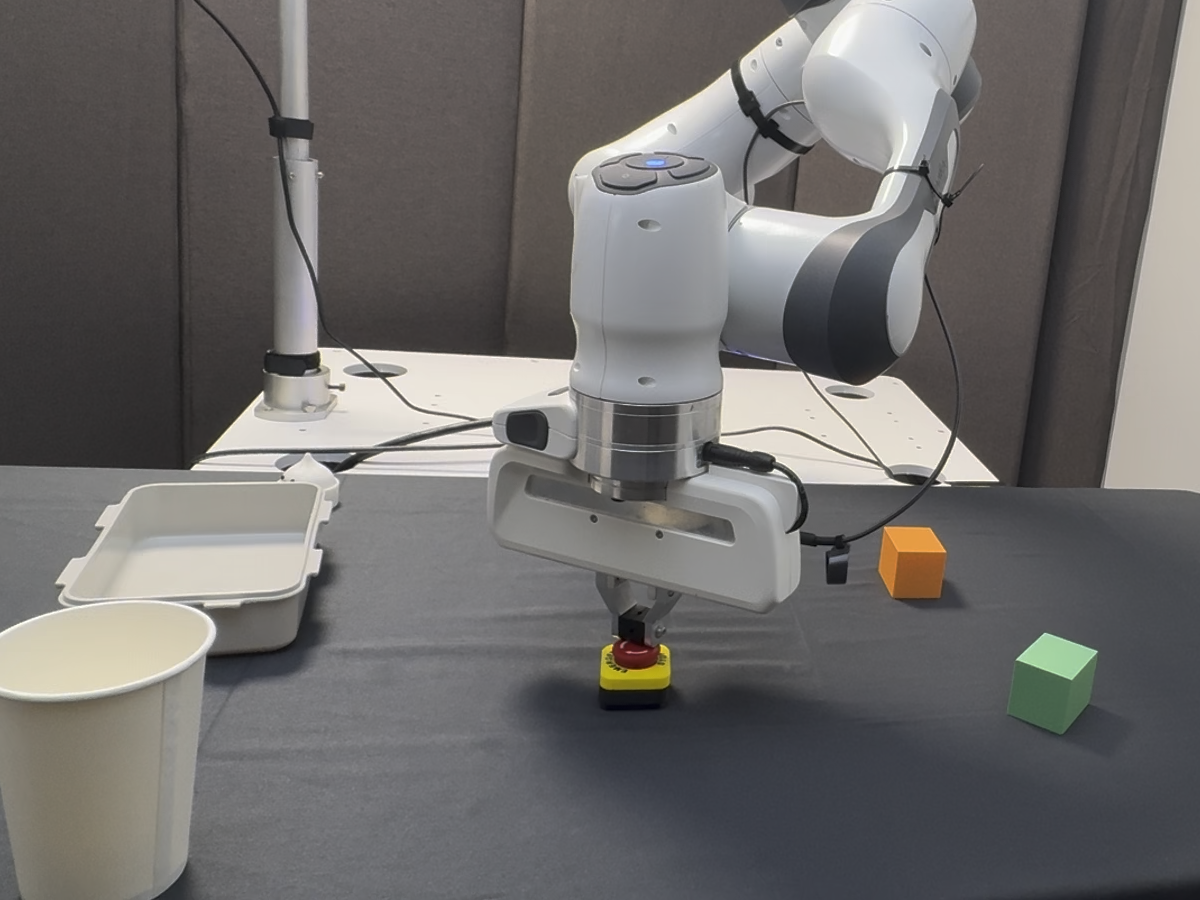}}
\par\smallskip
{\small (b) Franka Panda real robot}
\end{minipage}
\caption{\textbf{Simulation and real-robot evaluation environments.} (a) RoboLab scenes: food pickup, shelf placement, and bottle collection (left to right). (b) Franka Panda setups: Pick and Place, Stack Cube, and Press Button (left to right).}
\label{fig:evaluation-environments}
\end{figure}

%% file: sec/04_results.tex
\begin{table*}[t]
\centering\resulttablestyle
\caption{\textbf{Task success on RoboLab.} Language-agent methods use 120 episodes each; direct-action policies use 400. Success rates are reported overall and by task difficulty.}
\label{tab:robolab-main}
\vspace{6pt}
\setlength{\tabcolsep}{7pt}
\begin{tabular}{lrrrr}
\toprule
& \multicolumn{4}{c}{Task success (\%) $\uparrow$} \\
\cmidrule(lr){2-5}
Method & Overall & Simple & Moderate & Complex \\
& (40 tasks) & (21 tasks) & (13 tasks) & (6 tasks) \\
\midrule
\multicolumn{5}{l}{\emph{Direct-action policies}} \\
$\pi_{0.5}$ & 30.75 & 30.48 & 33.08 & 26.67 \\
Cosmos 3 & 42.25 & 44.76 & \textbf{45.38} & 26.67 \\
\midrule
\multicolumn{5}{l}{\emph{Code- and skill-based agents}} \\
mini-SWE & 20.83 & 25.40 & 17.95 & 11.11 \\
CaP-X & 38.33 & 42.86 & 35.90 & 27.78 \\
Open Robot Skill & 20.00 & 31.75 & 10.26 & 0.00 \\
\midrule
\multicolumn{5}{l}{\emph{Intent-based agent harness}} \\
\rowcolor{referenceblue!6}
\textbf{RobotUse} & \textbf{45.00} & \textbf{53.97} & 35.90 & \textbf{33.33} \\
\bottomrule
\end{tabular}
\vspace{-6pt}
\end{table*}

\section{Evaluation}
\label{sec:evaluation}

We evaluate whether RobotUse lets an agent carry out varied manipulation tasks while retaining control over how it acts. We then examine how context harnessing supports these decisions and how execution experience improves the guidance used in later attempts.

\subsection{Experimental Setup}
\label{sec:experimental-setup}

\paragraph{Environments.}
To evaluate generality across manipulation tasks, we use RoboLab~\citep{yang2026robolab}, whose tabletop tasks range from picking individual objects to arranging multiple objects under spatial and ordering constraints. We select 40 tasks spanning 21 Simple, 13 Moderate, and 6 Complex tasks, all with their default English instructions. To study improvement through physical execution, we use a Franka Emika Panda for Pick and Place, Stack Cube, and Press Button (Figure~\ref{fig:evaluation-environments}). Objects are randomly positioned within a $50 \times 50$\,cm workspace observed by a wrist camera and two fixed cameras.

\paragraph{Baselines.}
To compare how agents use robot capabilities, we include CaP-X~\citep{capx} and mini-SWE~\citep{yang2024sweagent}, which generate code, and Open Robot Skill~\citep{graphrobots2026openrobotskills}. Language agents use Gemini-3.8-Flash; mini-SWE shares the low-level backend used by CaP-X. We also include $\pi_{0.5}$~\citep{physicalintelligence2025pi05} and dense Cosmos 3~\citep{nvidia2026cosmos3} as direct-action policies on the same task catalog. The $\pi_{0.5}$ evaluation uses the public OpenPI checkpoint~\citep{physicalintelligence2025openpi,nvlabs2026robolabpi0}. Method-specific execution and stopping rules are given in Appendix~\ref{app:evaluation-details}.

\Needspace{7\baselineskip}
\paragraph{Metrics.}
We measure full-task success using the final task verifier. Language-agent results use three trials per task (120 episodes per method); direct-action policies use ten trials per task (400 episodes). The real-robot comparison uses 25 trials per method and task, with RobotUse's playbook frozen after a separate learning phase. For learning from experience, we examine performance across revisions together with the resulting changes in the playbook and subsequent execution.

\subsection{Task Performance}
\label{sec:main-results}

\paragraph{Overall Performance.}
We first compare task completion across the RoboLab catalog to assess how well the complete harness supports manipulation (Table~\ref{tab:robolab-main}). RobotUse succeeds in 54/120 episodes (45.00\%), exceeding CaP-X by 6.67 percentage points and Open Robot Skill by 25.00 points. It also exceeds mini-SWE (20.83\%), $\pi_{0.5}$ (30.75\%), and Cosmos 3 (42.25\%). RobotUse has the highest overall, Simple, and Complex success rates, while Cosmos 3 leads on Moderate tasks. Appendix Table~\ref{tab:robolab-attributes} reports the detailed task-attribute results.

%% file: sec/05_ablation.tex
\Needspace{19\baselineskip}
\begin{wrapfigure}{r}{0.48\textwidth}
\vspace{-12pt}
\centering
\includegraphics[width=\linewidth]{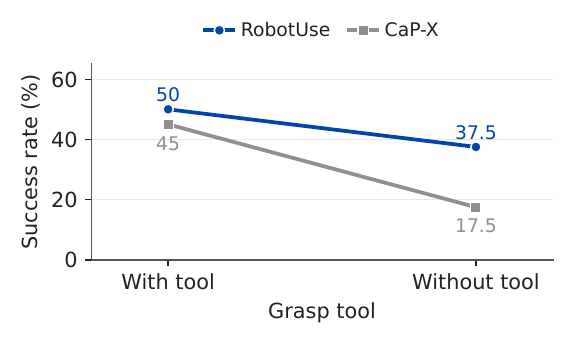}
\caption{\textbf{Robustness of RobotUse.} Success with and without the grasp tool; 40 tasks per~condition.}
\label{fig:flexibility}
\vspace{-8pt}
\end{wrapfigure}
\paragraph{Flexibility.}
\label{sec:ablation}
RobotUse lets the agent adjust its actions from visual feedback, supporting continued execution when tool outputs are imperfect or a tool is unavailable. Across 40 tasks, RobotUse retains 75\% of its original success rate after grasp-tool removal, falling from 50.00\% to 37.50\%; CaP-X falls from 45.00\% to 17.50\% (Figure~\ref{fig:flexibility}).

Figure~\ref{fig:qual}(a--g) follows the butter-box task from an identical initial scene. The segmentation mask covers only the top of the butter box. The agent uses the image to select an executable grasp, checks a measured 27\,mm offset and shifts the gripper by 20\,mm before closing, then raises the placement after three rejected paths. Each correction retains the selected target while revising how the gripper reaches or places it.

In the grasp-tool removal example on Stack3RubiksCube (Figure~\ref{fig:qual}(h--n)), the agent approaches a clicked point 8\,cm above the object. It then inspects the wrist image, lowers the gripper by 85\,mm, and rotates it by 38$^\circ$ to align the fingers with the cube faces. Later motion requests therefore reflect the inspected and adjusted pose. CaP-X's first two programs try to estimate cube orientation; one crashes and the other produces a rejected descent. Its remaining programs apply one fixed top-down orientation to every cube. The visual specification gives RobotUse a shared object for inspecting a choice, issuing a correction, and requesting the resulting motion.

%% file: sec/05_context.tex
\Needspace{19\baselineskip}
\subsection{Context Harnessing}
\label{sec:context-handoff}

\begin{wrapfigure}{r}{0.58\textwidth}
\vspace{-12pt}
\centering
\includegraphics[width=\linewidth]{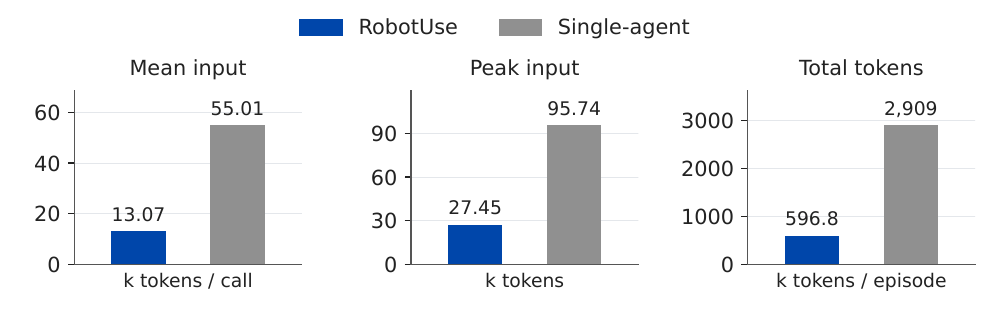}
\caption{\textbf{Context handoff.} Token usage for RobotUse and single-agent execution across 40 tasks and three seeds (120 episodes per condition). Mean and peak input are averaged across episodes; total tokens sum input and output within each~episode.}
\label{fig:context-usage}
\vspace{-8pt}
\end{wrapfigure}

Solving one subgoal can require several attempts, each adding observations, candidate selections, and diagnostic feedback. RobotUse retains this sequence in the local context used for spatial correction. A returned report carries the resulting scene and findings into the main agent's next task-level decision.

Fig.~\ref{fig:context-usage} shows the effect of context handoff across 120 episodes. We compare RobotUse with a single agent that performs all work otherwise handled by subagents, using the same tools and role instructions within one continuous conversation. Without subagents, mean input per call increases by $4.2\times$ and the average episode peak by $3.5\times$, raising total token usage by $4.9\times$ and API cost by $2.4\times$. Task success also drops from 45.0\% to 40.8\%. These results suggest that language reports from local visual interactions preserve sufficient context for subsequent decisions, while accumulating the full interaction history may burden execution. Appendix~\ref{app:shared-context} gives the execution protocol and Table~\ref{tab:context-ablation} reports the numerical results.

%% file: sec/05_continual.tex
\Needspace{4\baselineskip}
\subsection{Learning from Experience}
\label{sec:continual-evaluation}

\begin{wrapfigure}{r}{0.43\textwidth}
\vspace{-12pt}
\centering
\includegraphics[width=\linewidth]{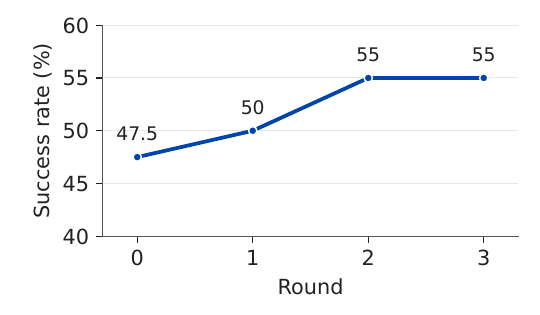}
\caption{\textbf{Learning from experience.} 40 tasks per revision.}
\label{fig:continual-curve}
\end{wrapfigure}

\begin{figure}[!htb]
\centering
\includegraphics[width=\linewidth]{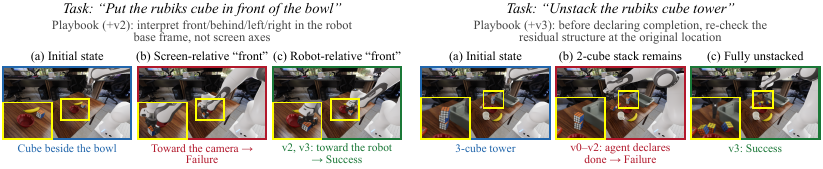}
\caption{\textbf{Qualitative playbook changes.} Left: robot-frame guidance grounds ``in front of the bowl.'' Right: checking the original stack enables complete unstacking.}
\label{fig:playbook-qualitative}
\end{figure}

\paragraph{Learning from Failures.}
Connecting the agent's choices to their observed outcomes makes execution failures informative: the agent can examine what went wrong and retain the resulting lessons in its playbook to guide subsequent choices. We trace this process across four harness revisions on the same 40-task RoboLab catalog. Losing track of targets during execution motivates a goal checklist. Misinterpreting spatial directions motivates guidance that defines them in the robot frame, while leaving cubes at the original stack motivates re-observation before declaring completion. These lessons change how the agent interprets goals and judges the results of its actions. Figure~\ref{fig:playbook-qualitative} illustrates the resulting choices when placing an object ``in front of the bowl'' and completing an unstacking task. Success increases from 19/40 to 22/40 across revisions (Figure~\ref{fig:continual-curve}); round 3 evaluates one frozen policy in 40 new episodes. Appendix~\ref{app:context-learning-results} details the revision protocol.

The revisions accumulate several kinds of guidance. A target checklist records the objects, counts, ordering, and relations that must remain satisfied throughout a task. Allowing tilt expands the pose adjustments available for a difficult grasp. Later guidance defines spatial terms in the robot frame and requires the agent to diagnose a rejected grasp before retrying it. The completion check returns attention to the original stack, while the orientation check asks the agent to confirm reorientation from the observed object. Each instruction connects an observed execution issue to a choice in a later interaction.

\par
\WFclear

\input{sec/04_real_robot}

%% file: sec/04_real_robot.tex
\Needspace{16\baselineskip}
\begin{wrapfigure}{r}{0.52\textwidth}
\vspace{-10pt}
\centering
\includegraphics[width=\linewidth]{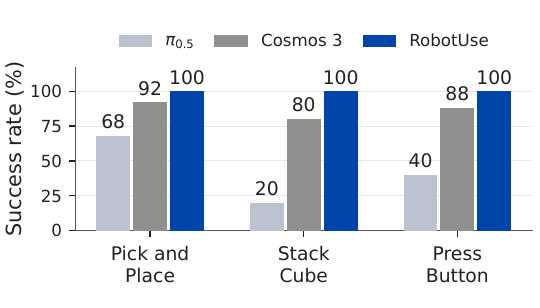}
\caption{\textbf{Real-robot task success.} Franka Panda; 25 evaluation trials per task. RobotUse freezes its playbook after its first learning success.}
\label{fig:real-robot-results}
\vspace{-8pt}
\end{wrapfigure}

\paragraph{Rapid Adaptation.}
On the Franka Panda, RobotUse adapts through playbook refinement while keeping model parameters and the execution backend fixed. We compare it with fine-tuned $\pi_{0.5}$ and Cosmos~3 baselines (Figure~\ref{fig:real-robot-results}). RobotUse first learns to pick up a cube and place it in a tray by reviewing each attempt and revising its playbook until the first success. The accumulated guidance also supports rapid adaptation to new tasks: carrying the playbook over to stacking a cube on another cube yields success on the first attempt. On the subsequent button-pressing task, only the first attempt fails, and the agent succeeds on its second attempt after further playbook refinement. With the playbook frozen at each task's first success, RobotUse succeeds in all 25 separate evaluation trials per task. These results show how experience retained in the playbook supports reliable execution and rapid adaptation to subsequent tasks without model fine-tuning.

%% file: sec/06_conclusion.tex
\section{Conclusion}
\label{sec:conclusion}

We introduced RobotUse, a robot agent harness that allocates computation, context, and decisions around specifying and revising physical actions. Visual action representations let agents express and revise how the robot should act, subgoal-local contexts retain the details needed for execution, and a persistent playbook carries experience into subsequent attempts. On RoboLab, RobotUse achieves 45\% task success and supports manipulation when the grasp tool is removed. Our context study shows lower token usage with subgoal delegation, while simulation and real-robot experiments illustrate how execution experience informs later decisions through playbook refinement. These findings support harness design as a means of improving robot agents: delegating geometric computation and control, preserving the physical choices agents need to revise, and retaining experience in guidance that shapes future action.

\newpage

%% file: sec/08_llm_statement.tex
\section*{AI Use Statement}
We used LLM-based assistants (Claude by Anthropic and ChatGPT by OpenAI) during the preparation of this work. Specifically, these tools were used to (i) polish the grammar, phrasing, and clarity of the manuscript text, and (ii) assist with writing and debugging code used in our experiments. All research ideas, experimental designs, and analyses were conceived and directed by the authors; LLM assistance was limited to implementation support carried out under our explicit instruction and supervision. The authors have reviewed and verified all AI-assisted content and take full responsibility for the contents of this paper.

%% file: sec/06_evaluation_details.tex
\section{Additional Evaluation Details}
\label{app:evaluation-details}

\subsection{Evaluation Protocols}

\paragraph{Tasks and protocol.}
We evaluate 40 RoboLab tasks with their default English instructions: 21 Simple, 13 Moderate, and 6 Complex tasks. The main comparison aggregates three runs indexed by seeds 0--2, with one episode per task in each run (120 episodes per method). Full-task success requires a successful final verifier result; all other episodes count as failures. Seed 0 uses fixed initial placements, while seeds 1 and 2 randomize dynamic-object positions independently within $\pm 0.10$\,m in X and Y.

\paragraph{Baselines and execution conditions.}
We compare RobotUse, CaP-X, mini-SWE, and Open Robot Skill. mini-SWE v1.17.5\footnote{\url{https://github.com/SWE-agent/mini-swe-agent/tree/v1.17.5}} uses the CaP-X M4 low-level interface with a persistent Python session. The completed language-agent reports use Gemini-3.8-Flash through OpenRouter for language-model inference. These are system-level comparisons with method-specific stopping rules. CaP-X uses the default set by its authors: an initial code execution and ten regenerations. RobotUse uses a 60-minute wall-clock limit per episode. Open Robot Skill uses a \$1.50 API-cost limit per episode. We set this limit because, without it, some Open Robot Skill episodes kept running in repeated loops and exceeded \$50 without succeeding, whereas successful episodes completed within \$1. The final API response may exceed this limit before further requests are stopped.

\paragraph{CaP-X motion on RoboLab.}
\label{app:capx-motion}
CaP-X's RoboLab integration reuses the simulator interface, joint controller, and inverse-kinematics solver of RobotUse's backend. For a fair comparison, its motion primitives apply the same path check as that backend: each commanded motion follows a straight-line Cartesian path interpolated at 3\,mm and 0.025\,rad, and is rejected when inverse kinematics fails or a joint changes by more than 0.2\,rad between waypoints. The upstream implementation instead moves in joint space to an inverse-kinematics solution. A rejected motion raises an exception in the generated program, which CaP-X can address when it regenerates code.

\paragraph{Direct-action reference.}
We additionally evaluate $\pi_{0.5}$~\citep{physicalintelligence2025pi05} using the public \texttt{pi05\_droid\_jointpos} checkpoint through the RoboLab-native OpenPI client~\citep{physicalintelligence2025openpi,nvlabs2026robolabpi0}. Each of the same 40 tasks runs in ten parallel environments, yielding 400 completed episodes with the default English instruction and native simulation limits. The runner uses native initialization and exposes no seed argument, so this reference uses a separate initialization and repetition protocol from the three language-agent runs. We aggregate the native binary success field across all episodes; the separately logged partial-task score is not used for success rates. $\pi_{0.5}$ achieves 123/400 successes (30.75\%): 64/210 Simple (30.48\%), 43/130 Moderate (33.08\%), and 16/60 Complex (26.67\%). Agent turns and external API cost are inapplicable to this local policy. Its batched timing records are kept separate from the per-episode agent process-time comparison.

\paragraph{Cosmos 3.}
The dense Cosmos 3~\citep{nvidia2026cosmos3} baseline evaluates each of the selected 40 tasks with ten rollouts, yielding 400 episodes. Success counts are 94/210 on Simple tasks, 59/130 on Moderate tasks, and 16/60 on Complex tasks, totaling 169/400.

\paragraph{Resource accounting.}
We average reported API cost, turns, tokens, and process time over all episodes. Effective tokens equal total tokens minus cached input tokens; $k$ denotes 1,000 tokens. Local perception-model and GPU costs are excluded. The main-agent comparison counts only RobotUse main requests and CaP-X code-generation requests, as defined below. Cost per success divides a run's total episode cost by its confirmed successes; Table~\ref{tab:resource-use} averages these ratios across the three runs. Process time measures per-episode elapsed time rather than parallel batch duration.

\subsection{Harness Revision Protocol}
This sequence measures iterative system development on the evaluated task catalog. All rounds use a 60-minute wall-clock limit per episode. Round 3 uses a single frozen policy. Evaluation costs exclude development probes and playbook-update costs.

\subsection{Grasp-Tool Comparison}
RobotUse uses a click location and relative height for the approach, with position and rotation adjustments, in its no-grasp run. The compared runs share seed 1 and randomized initial placements. These completed runs show a smaller observed success reduction for RobotUse.

\subsection{Context Handoff Ablation}
\label{app:shared-context}

\paragraph{Single-agent execution.}
The single-agent condition in Section~\ref{sec:context-handoff} uses the model, tools, role instructions, playbook v0, turn budgets, and execution backend of RobotUse. One agent performs all work otherwise handled by subagents within its own conversation. Each subgoal and its subsequent local interactions enter that conversation, and each role's instructions enter once, when the role is first used. Per-turn budgets and current images are supplied as in RobotUse. We evaluate all 40 tasks across three seeds (0--2), with one episode per task and seed (120 episodes per condition), using the existing RobotUse main-evaluation records for comparison. Seed 0 uses fixed placements, while seeds 1 and 2 randomize dynamic-object positions in X and Y as described above. The single-agent condition has a wall-clock limit of 60 minutes per task, with a 120-minute limit for CleanUpToysTask. Full-task success requires a successful final verifier result; failed and unverified endings remain in the denominator.

\paragraph{Accounting.}
For the input statistics in Table~\ref{tab:context-ablation}, we compute each episode's mean and maximum recorded input length, including cached input, then average each statistic equally across episodes. Total tokens include input and output over all model calls per episode, and API cost includes all agent roles. RobotUse uses 596.8k tokens and \$0.511 per episode, compared with 2,909k tokens and \$1.221 for single-agent execution. The average numbers of model calls per episode are 41.6 and 41.1, respectively. Success counts are 54/120 for RobotUse and 49/120 for single-agent execution.

\subsection{Main-Agent Turn and Context Accounting}
\label{app:main-context-accounting}
We recount the main agent from unique \texttt{agent\_input} events identified by session and step with role \texttt{prime}. A started decision with an unanswered request remains a turn; HTTP transport retries do not add turns.

Context length uses the provider's \texttt{prompt\_tokens} for each main-agent request, including cached inputs. We compute each episode's average and observed maximum over recorded usage, then average each statistic equally across its 40-episode campaign. Completion tokens and subagent calls are excluded from these context statistics but remain in the system-wide resource accounting.

For CaP-X seed 0, we read the archived \texttt{provider\_inputs.jsonl} and \texttt{provider\_audit.jsonl} files and count unique request identifiers with role \texttt{code}. This yields 259 main requests with complete input-token usage. VDM~\citep{capx} and Molmo~\citep{clark2026molmo2} calls are excluded from main-agent statistics. The cross-method main-agent comparison covers seed 0.

\begin{table}[t]
\centering\small
\caption{\textbf{Additional RobotUse main-agent accounting.} Each row contains 40 episodes. Definitions and coverage follow Table~\ref{tab:main-context}. Baseline seed 0 also serves as refinement round 0.}
\label{tab:robotuse-main-context-additional}
\setlength{\tabcolsep}{3pt}
\begin{tabular}{lrrrr}
\toprule
Run & Main turns/ep. & Mean input (k) & Peak input (k) & Coverage \\
\midrule
Baseline seed 0 / round 0 & 10.62 & 9.12 & 12.58 & 424/425 \\
Baseline seed 2 & 12.40 & 9.75 & 13.79 & 496/496 \\
Refinement round 1 & 11.08 & 10.01 & 13.64 & 443/443 \\
Refinement round 2 & 10.70 & 13.38 & 18.73 & 427/428 \\
Refinement round 3 & 11.25 & 14.37 & 20.13 & 450/450 \\
\bottomrule
\end{tabular}
\end{table}

\input{sec/09_context_learning_results}

\input{sec/08_additional_resources}

%% file: sec/09_context_learning_results.tex
\subsection{Context and Learning Results}
\label{app:context-learning-results}
Table~\ref{tab:context-ablation} gives the numerical context comparison in Section~\ref{sec:context-handoff}. Input lengths are computed per episode and then averaged across all 120 episodes. Tables~\ref{tab:continual-harnessing}--\ref{tab:playbook-analysis} retain the detailed results and revision records for Section~\ref{sec:continual-evaluation}.

\begin{table}[t]
\centering\resulttablestyle
\caption{\textbf{Context handoff ablation on RoboLab.} Forty tasks across seeds 0--2, one episode per task and seed (120 episodes per condition). Input/call is the mean input tokens per model call and Peak is the largest input to a single call, both computed per episode and averaged across episodes. Total tokens sum input and output over all calls per episode.}
\label{tab:context-ablation}
\setlength{\tabcolsep}{5pt}
\begin{tabular}{@{}lrrrr@{}}
\toprule
Configuration & Success & Mean input & Peak input & Total tokens \\
& (\%) $\uparrow$ & (k/call) & (k) & (k/ep.) \\
\midrule
\textbf{RobotUse} & \textbf{45.0} & \textbf{13.07} & \textbf{27.45} & \textbf{596.8} \\
Single-agent & 40.8 & 55.01 & 95.74 & 2909 \\
\bottomrule
\end{tabular}
\end{table}

\begin{table}[t]
\centering\resulttablestyle
\caption{\textbf{Continual harnessing on RoboLab.} Each round contains 40 seed-0 episodes. Cost includes all agent roles; turns count main-agent decisions. Protocol details appear in Appendix~\ref{app:evaluation-details}.}
\label{tab:continual-harnessing}
\setlength{\tabcolsep}{4pt}
\begin{tabular}{@{}lrrrrrr@{}}
\toprule
& \multicolumn{4}{c}{Success (\%) $\uparrow$} & \multicolumn{2}{c}{Usage per episode} \\
\cmidrule(lr){2-5}\cmidrule(lr){6-7}
Round & Overall & Simple & Moderate & Complex & Cost (\$) & Main turns \\
\midrule
0 & 47.50 & 61.90 & 30.77 & 33.33 & 0.4787 & 10.62 \\
1 & 50.00 & 66.67 & 38.46 & 16.67 & 0.5409 & 11.08 \\
2 & \textbf{55.00} & 71.43 & 46.15 & 16.67 & 0.5277 & 10.70 \\
3 & \textbf{55.00} & 76.19 & 30.77 & 33.33 & 0.5401 & 11.25 \\
\bottomrule
\end{tabular}
\end{table}

\begin{table}[t]
\centering\resulttablestyle
\caption{\textbf{Evaluation usage across harness revisions.} Per-episode token counts are in thousands. Cost/success divides evaluation costs by confirmed successes.}
\label{tab:continual-resources}
\begin{tabular}{@{}lrrrr@{}}
\toprule
& \multicolumn{2}{c}{Tokens per episode (k)} & Time (s/ep.) & Cost (\$/success) \\
\cmidrule(lr){2-3}
Round & Total & Effective & & \\
\midrule
0 & 555.69 & 502.59 & 1774.55 & 1.0077 \\
1 & 628.00 & 562.84 & 2082.74 & 1.0817 \\
2 & 632.88 & 552.59 & 1924.91 & 0.9594 \\
3 & 660.72 & 566.46 & 1785.98 & 0.9820 \\
\bottomrule
\end{tabular}
\end{table}

\begin{table*}[t]
\centering\resulttablestyle
\caption{\textbf{Playbook changes across harness revisions.} Representative additions (+), removals ($-$), and modifications (M).}
\label{tab:playbook-analysis}
\begin{tabular}{@{}l >{\raggedright\arraybackslash}p{4.3cm} >{\raggedright\arraybackslash}p{6.8cm}@{}}
\toprule
Version & Cause (observed issue) & Change \\
\midrule
v0 & --- & Baseline, brief skeletal guide. \\
\midrule
\multirow{2}{*}{v1}
 & Targets were lost or miscounted mid-task. & (+) Build a checklist of targets, counts, order, and relations. \\
\cmidrule{2-3}
 & Rotation restrictions blocked a needed pose. & ($-$) Remove the yaw-only constraint and allow tilt. \\
\midrule
\multirow{2}{*}{v2}
 & Screen and robot directions were confused. & (+) Define front/back/left/right in the robot frame. \\
\cmidrule{2-3}
 & Rejected grasps were retried unchanged. & (+) Diagnose the rejection before retrying. \\
\midrule
\multirow{2}{*}{v3}
 & Residual stacked objects were overlooked. & (+) Re-check the original location before declaring unstacking complete. \\
\cmidrule{2-3}
 & Gripper closure was mistaken for reorientation. & (M) Confirm orientation changes from observations. \\
\bottomrule
\end{tabular}
\end{table*}

%% file: sec/08_additional_resources.tex
\subsection{Additional Resource and Difficulty Breakdowns}
\label{app:additional-resources}
Effective tokens equal total tokens minus cached input tokens; $k$ denotes 1,000 tokens. Cost per success is the mean of run-level total-cost-to-success ratios. RobotUse uses more API cost and time than CaP-X and less API cost than Open Robot Skill.

\begin{table}[t]
\centering\small
\caption{\textbf{Additional inference usage.} Means over three runs; mini-SWE costs \$0.1706 per episode; token counts are in thousands per episode and time is seconds per episode. Cost/success is the mean of the three run-level ratios.}
\label{tab:resource-use}
\begin{tabular}{lrrrrr}
\toprule
Method & \$/ep. & Total tokens (k) & Effective (k) & Time (s) & \$/success \\
\midrule
CaP-X & 0.1672 & 119.94 & 119.20 & 642.80 & 0.4417 \\
Open Robot Skill & 1.3800 & 5779.41 & 1230.81 & 1185.77 & 7.1346 \\
\textbf{RobotUse} & 0.5110 & 596.80 & 555.33 & 1880.51 & 1.1984 \\
\bottomrule
\end{tabular}
\end{table}

\begin{figure}[t]
\centering
\includegraphics[width=.78\linewidth]{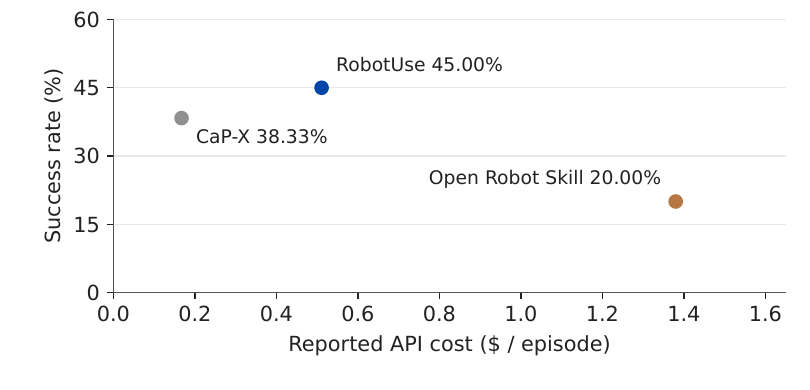}
\caption{\textbf{Task completion against reported API cost.} Each point summarizes the three reported runs of RobotUse, CaP-X, or Open Robot Skill. API cost averages all 120 episodes per method and excludes local perception and GPU costs. RobotUse attains higher success with higher reported cost than CaP-X and lower cost than Open Robot Skill.}
\label{fig:cost-success}
\end{figure}

\paragraph{Success by task attribute.}
Table~\ref{tab:robolab-attributes} breaks the success rates of Table~\ref{tab:robolab-main} down by RoboLab's task attributes, grouped into its visual, relational, and procedural categories. Among language/code agents, RobotUse has the highest or tied-highest success on nine of ten attributes; the exception is reorientation, which contains two tasks. Cosmos 3 leads on the three relational attributes and on semantics and affordance.

\begin{table}[t]
\centering\footnotesize
\caption{\textbf{Task success by RoboLab task attribute.} Success rates (\%) over the episodes of Table~\ref{tab:robolab-main}. Attributes and categories follow RoboLab's task metadata; a task can carry several attributes, and parentheses give the number of tasks.}
\label{tab:robolab-attributes}
\setlength{\tabcolsep}{2.6pt}
\begin{tabular}{@{}lrrrrrrrrrr@{}}
\toprule
& \multicolumn{3}{c}{Visual} & \multicolumn{3}{c}{Relational} & \multicolumn{4}{c}{Procedural} \\
\cmidrule(lr){2-4}\cmidrule(lr){5-7}\cmidrule(lr){8-11}
Method & Color & Sem. & Size & Conj. & Count. & Spatial & Afford. & Reor. & Sort. & Stack. \\
& (8) & (16) & (4) & (4) & (3) & (8) & (4) & (2) & (4) & (3) \\
\midrule
\multicolumn{11}{@{}l@{}}{\emph{Direct-action policies}} \\
$\pi_{0.5}$ & 21.25 & 26.88 & 55.00 & 50.00 & 83.33 & 13.75 & 17.50 & 25.00 & \textbf{22.50} & 16.67 \\
Cosmos 3 & 28.75 & \textbf{31.88} & 42.50 & \textbf{72.50} & \textbf{90.00} & \textbf{48.75} & \textbf{30.00} & 20.00 & \textbf{22.50} & 6.67 \\
\midrule
\multicolumn{11}{@{}l@{}}{\emph{Language agents}} \\
mini-SWE & 33.33 & 10.42 & 8.33 & 33.33 & 22.22 & 33.33 & 8.33 & 0.00 & 0.00 & 22.22 \\
CaP-X & 37.50 & 29.17 & 25.00 & 58.33 & 66.67 & 25.00 & 25.00 & \textbf{33.33} & 0.00 & \textbf{66.67} \\
Open Robot Skill & 16.67 & 6.25 & 33.33 & 58.33 & 22.22 & 16.67 & 0.00 & \textbf{33.33} & 0.00 & 0.00 \\
\addlinespace[2pt]
\rowcolor{referenceblue!6}
\textbf{RobotUse} & \textbf{41.67} & 31.25 & \textbf{66.67} & 58.33 & 88.89 & 33.33 & 25.00 & 16.67 & 16.67 & \textbf{66.67} \\
\bottomrule
\end{tabular}
\end{table}

\paragraph{Main-agent interaction and context.}
Table~\ref{tab:main-context} compares the main agent in RobotUse with the code-generating agent in CaP-X over the same 40 seed-0 tasks. RobotUse uses 10.62 main requests per episode and CaP-X uses 6.48. Their episode-averaged input contexts are 9.12k and 7.37k tokens per main request, respectively; average observed peaks are 12.58k and 11.90k. These statistics describe the input received directly by the main agent. CaP-X processes images through VDM and passes its descriptions to the code agent; the comparison therefore separates main-agent context burden from system-wide visual processing. Both runs use fixed initial placements and their recorded method-specific stopping rules. Appendix~\ref{app:main-context-accounting} gives the aggregation rules and additional RobotUse runs.

\begin{table}[t]
\centering\small
\caption{\textbf{Main-agent requests and input context on the 40 seed-0 tasks.} RobotUse counts role \texttt{prime}; CaP-X counts role \texttt{code}, excluding VDM and Molmo calls. Mean input averages per-episode mean input lengths; peak input averages observed per-episode maxima. Input lengths include cached tokens and use thousands of tokens. Coverage counts main requests with recorded input usage.}
\label{tab:main-context}
\setlength{\tabcolsep}{3pt}
\begin{tabular}{lrrrr}
\toprule
Method & Main turns/ep. & Mean input (k) & Peak input (k) & Coverage \\
\midrule
CaP-X & 6.48 & 7.37 & 11.90 & 259/259 \\
RobotUse & 10.62 & 9.12 & 12.58 & 424/425 \\
\bottomrule
\end{tabular}
\end{table}

\begin{table}[t]
\centering\small
\caption{\textbf{Performance with and without the grasp tool.} Each condition contains 40 seed-1 tasks. CaP-X success decreases by 27.50 percentage points and RobotUse by 12.50 points. RobotUse specifies no-grasp motions through a clicked location, height, and pose adjustments.}
\label{tab:no-grasp-detailed}
\setlength{\tabcolsep}{4pt}
\begin{tabular}{llrrrrr}
\toprule
Method & Grasp tool & Overall (\%) & Simple & Moderate & Complex & \$/ep. \\
\midrule
CaP-X & With & 45.00\% & 52.38\% & 38.46\% & 33.33\% & 0.1704 \\
CaP-X & Without & 17.50\% & 23.81\% & 7.69\% & 16.67\% & 0.2024 \\
RobotUse & With & 50.00\% & 47.62\% & 53.85\% & 50.00\% & 0.5704 \\
RobotUse & Without & 37.50\% & 47.62\% & 23.08\% & 33.33\% & 0.4100 \\
\bottomrule
\end{tabular}
\end{table}

\begin{table}[t]
\centering\small
\caption{\textbf{Resource use in the grasp-tool comparison.} Tokens include all agent roles and are thousands per episode.}
\label{tab:no-grasp-resources}
\setlength{\tabcolsep}{4pt}
\begin{tabular}{llrrrrr}
\toprule
Method & Grasp tool & Turns & Total (k) & Effective (k) & Time (s) & \$/success \\
\midrule
CaP-X & With & 13.70 & 126.43 & 125.52 & 622.63 & 0.3788 \\
CaP-X & Without & 15.75 & 145.66 & 143.09 & 963.83 & 1.1568 \\
RobotUse & With & --- & 641.37 & 584.98 & 1760.33 & 1.1407 \\
RobotUse & Without & --- & 420.13 & 370.87 & 1359.01 & 1.0933 \\
\bottomrule
\end{tabular}
\end{table}

%% file: sec/07_implementation.tex
\section{RobotUse Implementation}
\label{app:implementation}

\paragraph{Tool interface and spatial references.}
The implementation separates model-facing tool schemas from backend geometry and motion routines. A selected target is registered with the current observation and passed to subsequent tools by reference. The grasp and placement tools return candidate references, camera projections, and measured metadata. The agent inspects a candidate before selecting or editing it; the backend produces the detailed pose card on inspection and caches it for that candidate. An edit creates a revised candidate and its updated preview. Reference checks associate selections with the current interaction state, preventing an old candidate from being treated as a fresh executable selection after the scene changes.

\paragraph{Pose inspection and execution.}
Detailed cards combine side, top, and closing-plane views of measured points and the robot gripper mesh. Translation edits use robot-base XYZ axes in millimetres, while rotation edits use local gripper axes about the gripper contact center. Grasp and placement heights can be specified relative to a supported measured reference. The backend resolves the reference into a pose, checks the request, and computes the motion. Placement motion keeps the gripper closed, with release requested separately after observation. Tool feedback reports planning or execution status; task completion is determined by the native task verifier.

\paragraph{Context allocation.}
The runtime implements delegation with a fresh message list initialized from a role-scoped prompt and the delegated task. Pointing, grasp selection, placement, and local pose review use prompts and tool subsets for their respective operations. These prompt roles use the general-purpose language model specified in the evaluation setup. A child accumulates its own tool interactions; the parent receives the returned selection, status, reason, and relevant feedback rather than the full child transcript. Images and references are passed through the handoff and tool results. This implements the separation between task-level history and the detailed context used to make a spatial decision.

\paragraph{Playbook loading and revision records.}
The playbook is a versioned Markdown document with common guidance and role-specific sections. Each agent prompt receives the common section and the section for its current role. The loader records a SHA-256 hash of the document, and episode records preserve the selected version and configuration. Later revisions therefore change the guidance supplied to subsequent decisions while preserving earlier documents for comparison. The reported refinement campaign is detailed in Appendix~\ref{app:evaluation-details}.